# Towards a Precipitation Bias Corrector against Noise and Maldistribution

Xiaoyang Xu[1], Yiqun Liu[1], Hanqing Chao[1][0000−0001−5973−2343], Youcheng Luo[1], Hai Chu[2], Lei Chen[2], Junping Zhang[1][0000−−0002−5924−3360], and Leiming Ma[2]

[1] Shanghai Key Laboratory of Intelligent Information Processing,
School of Computer Science, Fudan University, Shanghai 200433, China
{xuxy17,yqliu17,hqchao16,luoyc18,jpzhang}@fudan.edu.cn
[2] Shanghai Central Meteorological Observatory, China
{chhaui,qqydss}@163.com, {malm}@typhoon.org.cn

**Abstract.** With broad applications in various public services like aviation management and urban disaster warning, numerical precipitation prediction plays a crucial role in weather forecast. However, constrained by the limitation of observation and conventional meteorological models, the numerical precipitation predictions are often highly biased. To correct this bias, classical correction methods heavily depend on profound experts who have knowledge in aerodynamics, thermodynamics and meteorology. As precipitation can be influenced by countless factors, however, the performances of these expert-driven methods can drop drastically when some un-modeled factors change. To address this issue, this paper presents a data-driven deep learning model which mainly includes two blocks, i.e. a Denoising Autoencoder Block and an Ordinal Regression Block. To the best of our knowledge, it is the first expert-free models for bias correction. The proposed model can effectively correct the numerical precipitation prediction based on 37 basic meteorological data from European Centre for Medium-Range Weather Forecasts (ECMWF). Experiments indicate that compared with several classical machine learning algorithms and deep learning models, our method achieves the best correcting performance and meteorological index, namely the threat scores (TS), obtaining satisfactory visualization effect.

## 1 Introduction

Weather Forecast is an essential public service for everyone's life because of its wide applications in all aspects of society. For instance, weather warning is significant to protect life and property. Temperature and precipitation forecasts are important to agriculture and urban management. The earliest numerical weather prediction (NWP) can be dated back to Charney et al.'s work [2] that is based on atmospheric dynamics in 1950. From then on, NWP gradually becomes the dominated method of weather forecast. Different from the conventional subjective weather forecast methods that depend on well-trained meteorological experts, NWP estimates the state of the fluid at some time in the future by analyzing the state of the fluid at a given time quantitatively with aerodynamic and thermodynamic models. Due to errors from observations and the mismatching issue stemming from the nature of the numerical methods [7], however, NWP results often contain considerable bias. To correct the bias, many bias correction methods have been developed, such as Model Output Statistics (MOS) [15].

For weather forecast of and beyond 24 hours [3], NWP methods have achieved satisfactory accuracy with existing bias correction techniques. For the forecast within 6 hours, however, subjective judgements depending on expert forecasters are still main way for correcting NWP results because too many unknown and complex factors are involved.

Different from the weather prediction task taking observational results as inputs, the inputs for bias correction task are highly noisy NWP results. These noises are mainly introduced by two ways, i.e. errors from observations and errors from NWP models. Besides, the highly imbalanced data distribution and a large span between the maximum and minimum precipitations make the bias correction task even more challenging. For example, nearly 60% samples in our data are non-precipitation, and only in 1% samples the precipitation value is larger than 10. Meanwhile, the maximum precipitation value is 151, which is 1500 times of minimum precipitation value 0.1.

To address these issue mentioned above, we propose a novel deep convolution framework named Ordinal Boosting AutoEncoder (OBA) for the bias correction of numerical precipitation prediction. And we also finely design our model structure and the loss function. First, to denoise the NWP results, an Enhanced Denoised Autoencoder (EDA) is deployed to extract noise-free features from the highly noisy inputs. To get more robust features, EDA uses a Perturbation Layer that directly introduces noise into the intermediate feature of the Encoder block. Second, to conquer the maldistribution and long span of data, an Ordinal Boosting Regression block is employed to generate corrected results based on the extracted denoised features. In Ordinal Boosting Regression block, the regression task is changed into multiple ordinal binary classification tasks, so that different binary classification tasks can focus on the correction of different precipitation level. For instance, one of the tasks is focused on correcting samples ranging from 20.0mm to 20.5mm while another task is focused on 0.5mm to 1.0mm. Compared with a conventional regression structure, Ordinal Boosting Regression block is superior in tasks with long-span data. What's more, Focal

---

[3] According to the prediction time range, weather prediction can be categorized into nowcasting (0-2 hours), very short-range weather forecasting (0-12 hours), short-range weather forecasting (12-72 hours), Medium-range weather forecasting (72-240 hours), extended-range weather forecasting(10-30 days) and long-range forecasting(>30 days)

Loss [14] is applied to alleviate the influence of the maldistribution. The main contributions of this paper are summarized as follows:

1. By introducing deep learning structures into the bias correction of numerical weather prediction, our method achieves better correcting performance and meteorological index compared with several baselines;
2. With a perturbation layer, the proposed Enhanced Denoising Autoencoder can encode more robust features in the condition of higher noise perturbation;
3. Ordinal Boosting Regression Block is presented to eliminate the impact of the large span and focal loss is introduced in this block to address the issue of long-tail distribution.

## 2   Related Work

### 2.1   Bias Correction of Numerical Weather Prediction

The performance of NWP models is easily degenerated by model bias and inaccuracies included in initial condition, as indicated in [24]. The goal of the bias correction of NWP is to use the biased data generated by the NWP models to predict the precise weather indicators. The conventional approaches for bias correction are subjective and expert-driven. Thus, the performance of these subjective methods is limited by the modeled factors. Recently, some objective approaches [8][1] are proposed for post-processing the predicted parameters of any NWP models and handle the prediction bias in view of statistics or meteorology. Among them, Model Output Statistics (MOS) [7] is the most practical bias correction method and it typically generates multiple predictors by multiple linear regression. For the bias correction of precipitation, MOS techniques have been used to develop specific algorithms to forecast the probability of precipitation and rain amount in Australia [23] and Spain [25]. However, the predictors of MOS are based upon the three-dimensional fields produced by NWP models, surface observations and the climatological conditions for specific locations which might not be not applicable to obtain surface observations at some location. Therefore, the results of the MOS are valid only in limited areas. Apart from using the single model to correct the biased prediction, efforts also have been paid to verifying the ensemble model [4]. Bayesian Model Averaging (BMA) [19] is derived from the Bayesian probability decision theory and revise the ensemble forecast of rainstorm probability [26]. Although BMA improves the correction performance at the low-graded precipitation, it still mainly bases on the former observational data and the prediction of BMA is much too coarse with the grid distance of 50 km. All of the works mentioned above propose some handcrafted features based on prior knowledge and expert experience first, and then adopt linear models to fuse the observational and hand-crafted features to correct bias. To the best of our knowledge, there is no expert-free algorithm such as deep learning being introducing for the bias correction of NWP.

## 2.2   Ordinal Regression

Ordinal Regression aim at finding a function to predict the labels in ordinal variables. Most ordinal regression algorithms are derived from the classical classification algorithms. For instances, Herbrich et al. [10] proposed a support vector learning method for ordinal regression. Crammer et al. [3] proposed the perceptron ranking (PRank) algorithm to predict ordinal labels by generalizing the online perceptron algorithm with multiple thresholds. Shashua and Levin [21] developed a new support vector machine to handle multiple thresholds.

Another perspective for ordinal regression is regarding it as a series of binary classification problems. Frank et al. [5] adopted several decision trees for binary classifiers for ordinal regression. Niu et al. [18] combined ordinal regression with Convolution Neural Networks to address age estimation and Fu et al. [6] adopted a similar strategy for monocular depth estimation.

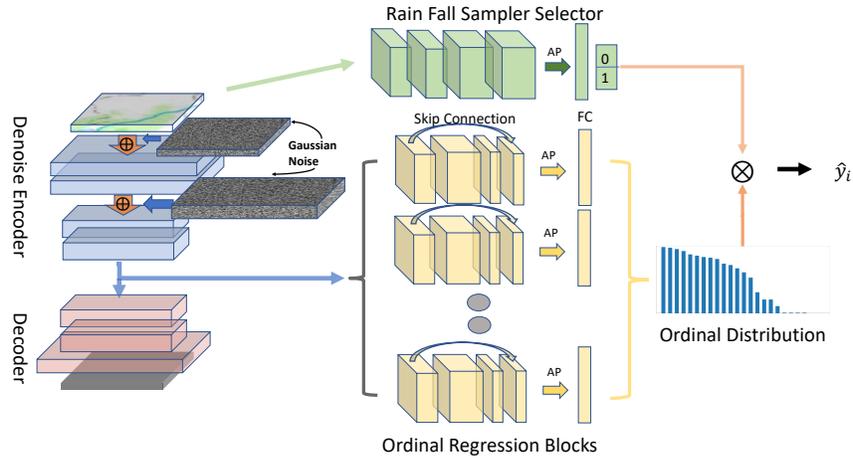

**Fig. 1.** An overview of proposed approach for the bias correction of numerical precipitation prediction. 'AP' represents Average Pooling and 'FC' is the abbreviation of the Fully Connection Layers. Cubes represent input data or feature maps of convolution blocks in our model and rectangles is the output vectors of FCs. The output vectors of the ordinal regression blocks form the ordinal distribution histogram. The final predicted values are generated by the one-hot vectors of the rainfall sampler selector.

## 3   Proposed Framework

In this section, we present ordinal boosting autoencoder for the bias correction of numerical precipitation prediction. We firstly describe the definition of this task, followed by the design for handling the noised features. For the purpose of easing the impact of the large span and imbalance distribution of precipitation values, an ordinal boosting regression block is implemented and details of this are described in Section 3.3.

### 3.1   Problem Definition

In this paper, we regard the bias correction of numerical precipitation prediction as a regression problem. Specifically, let $\mathcal{X}$ and $\mathcal{Y}$ denote the input and output spaces, respectively. Each input sample $x \in \mathcal{X}$ consists of $N$ features $f_1, f_2, \cdots, f_N$, in which the $f_i$ is a 2-dimensional matrix, with each element recording a value of the $i$-th feature on the corresponding latitude and longitude, and each output value $y \in \mathcal{Y}$ is a scalar representing for the corrected precipitation value.

All features are the multiple predicted meteorological parameters generated by the European Centre for Medium-Range Weather Forecasts. Among them, one is the biased precipitation values to be corrected. Therefore, our goal becomes to learn a mapping function $g : x \to y$. Note that, not only the features contains random noise but also precipitation values own a large span and long tail distribution. This makes the biased correction task become extremely challenging.

To solve these issues, we thus proposed a deep learning framework to obtain a better solution. The framework includes three parts (see Fig 1): an Autoencoder for robust features extraction, a Rainfall Sampler Selector for the dataset balancing and multiple Ordinal Regression Blocks for ordinal distribution learning.

### 3.2   Dealing with Noised Features

As mentioned above, input features with observational and systematical errors lead regression model to generate a precipitation value with bias. To issue this problem, we refine a Denoising Autoencoder with fully convolutional neural network through introducing noisy level. Formally, we can formulate it as follows:

$$z_i = g_{E_1}\left(x_i + \varepsilon \parallel \theta\right) \tag{1}$$

$$h_i = g_{E_2}\left(z_i + \varepsilon \parallel \theta'\right) \tag{2}$$

$$\hat{x}_i = g_D\left(h_i \parallel \theta''\right) \tag{3}$$

where $\varepsilon \sim N\left(\mu, \sigma\right)$ is a random normal noise. $g_{E_i}\left(\cdot\right)$ denotes each part of enhanced denoising encoder, $g_D\left(\cdot\right)$ denotes enhanced denoising decoder and $\theta, \theta', \theta''$ denote their respective parameters. $z_i$ is the output feature map of the layer ahead of the perturbation layer. $h_i$ is the bottleneck output of Denoising Autoencoder. With such a setting, the objective function of the Denoising Encoder-Decoder can be formulated as

$$\theta, \theta', \theta'' = \underset{\theta, \theta', \theta''}{\arg\min} \sum_i^N E_x \left\| x_i - \hat{x} \right\|_2^2 \tag{4}$$

During the test, we take original samples $x_i$ as inputs and remove perturbation layer and the $z_i$ can be regarded as noise-free and squeezed features, which will be utilized in ordinal regression block to correct the precipitation values.

### 3.3   Ordinal Boosting Regression Block

As a regression problem, one common objective function is least square error loss:

$$\mathcal{L}_R = \frac{1}{N} \sum_{i=1}^{N} (y_i - \hat{y}_i)^2 \tag{5}$$

however the ground-truth precipitation values (See Fig. 2) suffer the large span and the distribution of them is typical long-tail distribution. Therefore, learning such a distribution with equation (5) will force the regression model to favor some small precipitation values while the enormous precipitation values still bring large loss values, which causes model hardness to converge to the global optimum.

Furthermore, the precipitation value is a typical ordinal variable, whose value exists a significant ordering relation on arbitrary scale. Therefore, we adopt multiple ordinal boosting regression blocks to predict precise precipitation values. Concretely, we transfer an ordinal regression problem to a series of binary classification problems like [18]. In our work, we partition the 6-Hours cumulative precipitation values ranging from $y_{\min}$ to $y_{\max}$ into $K$ ranks $r_i \in \{r_1, r_2, \cdots, r_{K-1}\}$ followed by training $K-1$ binary classifiers with an interval 0.5 to judge whether the rank of the predicted values of the sample $x_i$ is larger than $r_k$.

More concretely, we firstly transform the continuous ground-truth precipitation value $y_i$ to discrete binary vectors $\mathbf{y}_i = \{d_1, d_2, \cdots, d_{K-1}\}$, where $d_i$ is calculated as follows:

$$d_i = \begin{cases} 1, \text{if} \quad y_i > r_k \\ 0, \text{ otherwise,} \end{cases} \tag{6}$$

Then, we directly utilize $K-1$ four-layers CNN as the boosting subnets, each of which generates a scalar $p_i$ presenting the probability of $d_i = 1$. The details of each boosting subnet will be shown in Section 4.3. Lastly, the ordinal boosting regression blocks calculate the predicted precipitation value of the sample $x_i$ as follows:

$$\tilde{y}_i = \eta * \sum_{i}^{K-1} (p_i \geq \xi) \tag{7}$$

where $\eta$ is the partitioning interval 0.5 and $\xi$ is a threshold set as 0.5 in our experiment. In such a setting, the samples with larger precipitation values contribute equally to the ordinal regression model with those with smaller values. As a result, the impact of the large span issue can be alleviated. However, the partitioning interval is larger than the minimal ranging interval in some cases resulting in inevitable gaps between the predicted values and the ground-truth values. Therefore, the partitioning interval is a very important hyper-parameter for our approach and we discuss more in Section. 5.2

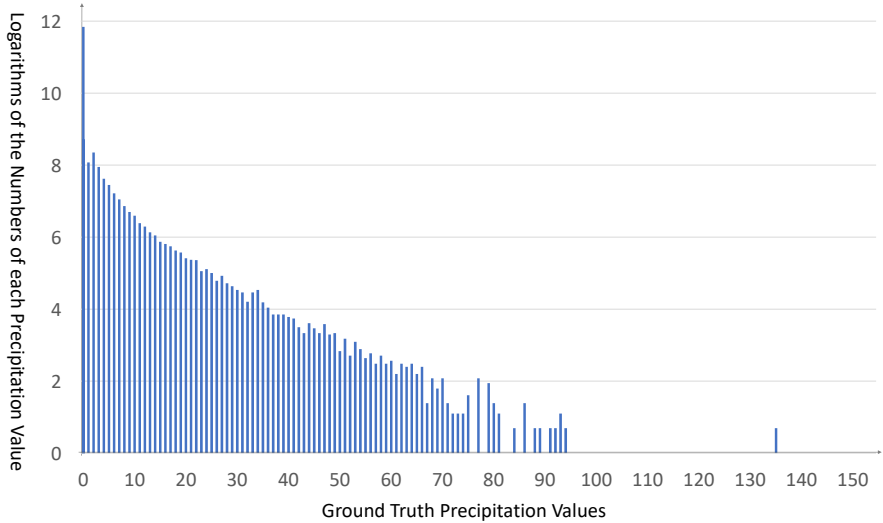

**Fig. 2.** The ground-truth precipitation values distribution on our dataset. The y-axis represents the logarithms of the numbers of each precipitation value and the x-axis represents the range of the ground-truth precipitation values.

### 3.4    Handling Imbalance Data

As shown in Figure 2, it is worthy noting that the majority of the ground-truth precipitation values of samples are fewer than 10 and the distribution of those is a long-tail distribution. Such a long-tail distribution further causes the ordinal ranked-classes $r_i$ imbalance, degenerating the predictive performance as most of samples belonging to negative won't bring any efficient signals. Focal Loss [14] is adopted in our approach to address imbalance, which stems from the cross-entropy loss function and designs specially for balancing the data and differentiating the easy/hard samples to make the harder ones contribute more. We combine focal loss with the loss function of ordinal boosting regression blocks to address the issues of imbalance categories and the loss function can been seen as a variant of cross-entropy loss function:

$$\mathcal{L}_{OR} = \frac{1}{N} \sum_{i=1}^{N} \mathbf{y}_i \log(1 - \mathbf{p}_i) + (1 - \mathbf{y}_i) \log(\mathbf{p}_i) \tag{8}$$

where both $\mathbf{y}_i$ and $\mathbf{p}_i$ are vectors representing for the labels and probabilities of each ordinal rank class $r_i$, respectively. Same as [14], we introduce two hyper-parameters in $\mathcal{L}_{OR}$ and rewrite it as:

$$\mathcal{L}_{FOR} = \frac{1}{N} \sum_{i=1}^{N} (1 - \boldsymbol{\alpha}) \mathbf{p}_i^{\gamma} \mathbf{y}_i \log(1 - \mathbf{p}_i) + \boldsymbol{\alpha}(1 - \mathbf{p}_i)^{\gamma} (1 - \mathbf{y}_i) \log(\mathbf{p}_i) \tag{9}$$

where $\gamma$ is a scalar for differentiating the samples and $\boldsymbol{\alpha}$ is a vector to balances each $r_i$ rank class.

Another problem occurred by the imbalance data is that the amount of the non-precipitation samples with label of 0 is much larger than the minimum precipitation samples with label of 0.1 while the loss between two of them calculated by Equation (5) is tiny. As a result of this problem, our model tends to predict the samples with minimum precipitation as non-precipitation samples, which is a serious mistake even the errors are fewer than 0.1. To address this issue, we adopt a strategy that we only feed the rainfall samples to OBA when training and individually train a rainfall samples selector for predicting if this sample is a precipitation sample when testing.

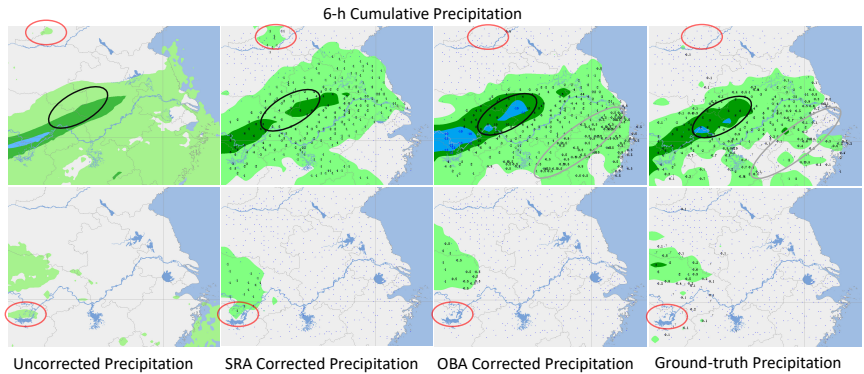

**Fig. 3.** Two visualization examples of our correction results. Uncorrected Precipitation (Left 1) shows the raw European Model precipitation grid data. SRA Corrected Precipitation (Left 2) represents corrected precipitation values of SRA. ORA Corrected Precipitation (Right 2) is corrected precipitation values of ORA. Ground-truth Precipitation (Right 1) is the ground-truth precipitation values.

## 4   Experiments

In this section, we first introduce our datasets and pre-processing we used, then we describe the evaluation metrics followed by model configurations and training details, and finally we report empirical experiments to validate the effectiveness of OBA in bias correction task.

### 4.1   Data Source

We evaluate OBA on 6-Hours Integrated Forecasting System (IFS) data [17] provided by the European Centre for Medium-Range Weather Forecasts (ECMWF).

These data present more than 670 meteorological parameters with different altitudes and all of them are the predictive values by some model of the NWP. These parameters are filled in a chart of world with latitude-longitude grid cells of 0.125° latitude by 0.125° longitude. Due to a combination of the observation bias causing by limited instrument accuracy and the systematic NWP model error, these predictive values contain lots of biases to be corrected. As comparison, the precise precipitation values come from the ground observatories. As the geographical coordinates of the ground observatories mismatch those of the grid data, an important hypothesis is proposed that the precise precipitation value inside a grid cell are uniform.

For the purpose of the correction the precipitation prediction bias, we firstly screen out some meteor parameters according to the ranks of correlation with precipitation, which are calculated with Pearson Correlation CoefficientP, as the input features for our bias correction model. The picking threshold is set as 0.2 and parameters with the same name but different altitudes only maintain the one with highest coefficient. Then, considering the spatial impact of the rainfall, we slice the entire grid matrix to several smaller ones centering at the ground observatory with the slice window of 2.0° latitude by 2.0° longitude. After preprocessing the entire grid, every sample for the model is resized to $37 * 17 * 17$. In our experiment, the whole data set consist of 237,498 samples of Ease China collecting from July to September in both 2016 and 2017. And the ratio of training set and test set is 4:1, and samples are randomly selected without overlapping.

### 4.2 Evaluation Metrics

We report the mean absolute error (MAE), mean precipitation absolute error (MPAE), and $Ts_\delta$ Score. MAE is defined as the mean absolute error between the ground precipitation values and the corrected precipitation values, and MPAE is the MAE excluding the cases without any rainfall. Owing to the limited measurement range of instruments, the minimum precipitation value is set as 0.1.

The Threat Score (Ts) is an important meteorological index with a parameter $\delta$ for NWP and is calculated as: $Ts_\delta = \frac{TP_\delta}{TP_\delta + FP_\delta + FN_\delta}$, where $\delta$ defines the boundary of positive samples and negative samples. Specifically, samples with label larger than $\delta$ are regarded as positive samples and those with label smaller than $\delta$ are negative samples.

### 4.3 Models Configurations and Training Details

Details of models configurations are shown in Table 1 and each Conv Block is equipped with a convolution layer, a batch normalization layer [11] and a leaky ReLU with a negative slope of 0.01 [16].

The denoising encoder model contains four Conv-Blocks and one noise perturbation layer is added in the middle of the encoder. We sample random noise from a normal distribution with $\mu = 0$ and $\sigma = 0.01$ when training and remove them during testing phase. The denoising decoder consists of three up-sampling Conv-Blocks, whose sampling operation is bilinear interpolation with stride 2.

**Table 1.** The Model Details of the Proposed Framework

| | | | |
|---|---|---|---|
| **Input**: $37 \times 17 \times 17$ | | | |
| **Encoder** | Conv Block | $\begin{bmatrix} 32, \ 1*1 \\ 32, \ 3*3 \end{bmatrix} \times 1$ | |
| | Noise Perturbation Layer | $N(0, 1e-3)$ | |
| | Conv Block | $\begin{bmatrix} 64, \ 3*3 \end{bmatrix} \times 2$ | |
| **Decoder** | Conv Block | $\begin{bmatrix} 64, \ 3*3 \end{bmatrix} \times 1$ | |
| | Upsampling | Bilinear $(17 \times 17)$ | |
| | Conv Block | $\begin{bmatrix} 32, \ 3*3 \\ 37, \ 1*1 \end{bmatrix} \times 1$ | |
| **Ordinal Subnets** | Conv Block | $\begin{bmatrix} 128, \ 3*3 \\ 128, \ 3*3 \\ 32, \ 1*1 \\ 32, \ 1*1 \end{bmatrix} \times 1$ | |
| | Down Sampler | $\begin{bmatrix} 32, \ 1*1 \end{bmatrix} \times 1$ | |
| | Binary Classifier | Average pooling, 1-D FC | |
| **RainFall Sampler** | Conv Block | $\begin{bmatrix} 64, \ 3*3 \\ 64, \ 3*3 \\ 128, \ 3*3 \\ 128, \ 3*3 \end{bmatrix} \times 1$ | |
| | Down Sampler | $\begin{bmatrix} 128, \ 1*1 \end{bmatrix} \times 1$ | |
| | RainFall Classifier | Average pooling, 1-D FC | |

Although all of the ordinal boosting subnets share one configuration, their parameters are completely independent. Each ordinal subnet contains one skip connection [9] to accelerate the training procedure and dropout [22] with a ratio of 0.2 is used for relieving the overfitting of the Full Connection Layers (FCs).

We adopt features normalization before features are fed into the models and use the Adam [12] with a mini-batch of 256. The learning rate of the denoising model starts from $1e-3$ with a weight decay of $1e-2$. Both the learning rate and weight decay of the ordinal boosting subnets is set as $1e-4$.

In testing, for comparison with other methods, we adopt the standard 5-fold cross validation. Noted that all experiments in Section 5 are implemented with a fixed dataset of 80% for training and 20% for testing.

### 4.4 Validation on 6−Hours IFS

We validate our ordinal boosting autoencoder on a 6−Hours IFS dataset by comparison with several baseline methods. Bilinear Interpolation (BI) means that we directly interpolate each sliced predicted precipitation grid to the corresponding ground observatory. Linear Regression (LR) and Support Vector Regression (SVR) take all interpolated features as inputs and directly predict the corrected precipitation values. We also implement a Multilayer Perceptron (MLP) [20] and a simple Fully Convolution Network [13] with four Conv-Blocks as two baseline deep learning models. For the validation of ordinal regression, we deploy a Single Regression Autoencoder (SRA) by replacing ordinal boosting blocks with a single regression neural network that has the same structure as an ordinal regression subnet.

**Table 2.** Comparison of our proposed model against baseline method on 6−Hours IFS

| Methods | MAE | MPAE | $Ts_{0.1}$ | $Ts_1$ | $Ts_{10}$ |
|---|---|---|---|---|---|
| BI | $1.31 \pm 0.08$ | $4.52 \pm 0.32$ | $0.44 \pm 0.03$ | $0.43 \pm 0.03$ | $0.24 \pm 0.02$ |
| LR | $1.60 \pm 0.05$ | $4.43 \pm 0.21$ | $0.31 \pm 0.02$ | $0.35 \pm 0.04$ | $0.20 \pm 0.02$ |
| SVR | $1.34 \pm 0.05$ | $4.96 \pm 0.24$ | $0.23 \pm 0.01$ | $0.35 \pm 0.03$ | $0.00 \pm 0.00$ |
| MLP | $1.25 \pm 0.04$ | $4.25 \pm 0.15$ | $0.40 \pm 0.03$ | $0.45 \pm 0.02$ | $0.29 \pm 0.02$ |
| FCN | $1.25 \pm 0.11$ | $4.36 \pm 0.28$ | $0.39 \pm 0.04$ | $0.45 \pm 0.05$ | $0.29 \pm 0.03$ |
| SRA | $1.20 \pm 0.10$ | $4.30 \pm 0.12$ | $0.58 \pm 0.03$ | $0.47 \pm 0.02$ | $0.29 \pm 0.01$ |
| OBA (ours) | $\mathbf{1.02 \pm 0.03}$ | $\mathbf{4.23 \pm 0.14}$ | $\mathbf{0.60 \pm 0.03}$ | $\mathbf{0.52 \pm 0.01}$ | $\mathbf{0.30 \pm 0.01}$ |

We summarize the results of OBA and several competitors in Table 2 and present the result by 'mean± standard deviation' on five metrics. As can be seen, OBA outperforms others on all metrics. Note that, our approach is the only method that has a $Ts_{0.1}$ above 0.6 and a MAE below 1.05. It is obvious that avoiding the ordinal regression blocks learning on non-precipitation samples helps the proposed OBA to gain a better performance on distinguishing tiny precipitation from non-precipitation. We visualize several examples of our precipitation bias correction results, shown in Figure 3.

Specifically, in Figure 3 the red circles show the cases in which our OBA model correctly revises the non-precipitation region. The black circles show the regions where OBA successfully correct the predictions with small rainfall into large rainfall. In addition, the gray circles show a more interesting circumstances. In the figure of ground truth, we can see that there are multiple areas in the gray circles have small rainfall which means that the whole region is in a critical state of raining. In such a region, the uncorrected prediction obviously overestimates the area of the region and the prediction of SRA over-estimates the rain amount in the region. Visualization examples show that OBA owns the ability of rectifying the biased precipitation values and rearranging the non-precipitation and precipitation values. Almost all the predictions of OBA are 0.5mm. The reason why OBA does not generate smaller predictions is that the most fine-grained of the Ordinal Boosting Regression is 0.5mm. Thus when the Rainfall Sampler Selector indicates the region has a high possibility of raining, the minimal prediction that OBA can generate is 0.5mm. This limitation could raise some indices like MAE or MSE but in real-life application since the region is indeed have a high possibility to rain, such prediction is still acceptable.

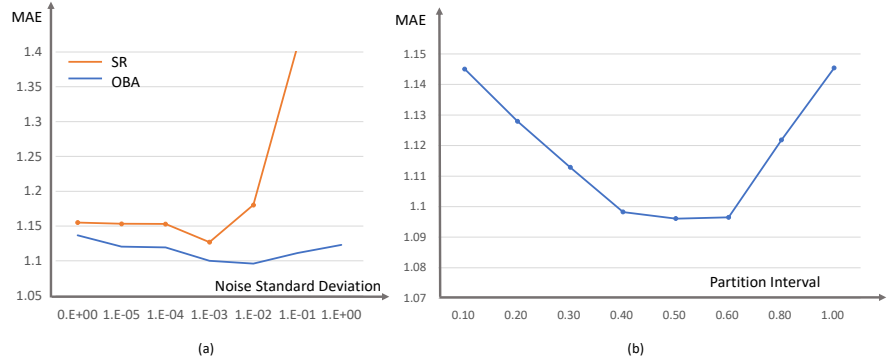

**Fig. 4.** Performance changes by varying (a) two types of standard deviation and (b) partition interval.

## 5   Discussion

### 5.1   Ablation Study

We analyze the effectiveness of two novel parts of our approach by ablation experiments and the results are shown in Figure 4(a).

**The Impact of Ordinal Regression** The orange line in Figure 4(a) represents the results of single regression model and the blue one represents the results of OBA. As indicated in the comparison of two lines, the MAE of using ordinal regression is always less than that of the single regression models. The main reason for this phenomenal improvement is that converting a regression problem to multiple classification problems alleviates the impact of the large span of the training labels.

**Influence of the Noise Perturbation** For analyzing the influence of the noise perturbation, we change the standard deviation $\sigma$ of the normal distribution. As shown in Figure 4(a), the MAE of our OBA goes down first and then up with the increase of $\sigma$. When the $\sigma$ is set as 0, the denoising autoencoder is degraded to a simple autoencoder and the features of its bottleneck layer are squeezed without noise handling. Therefore, the features with noise lead to the worst performance among all of the $\sigma$ setting. With the increase of $\sigma$, the denoising autoencoder starts to learn how to deal with the noise and the performance of our model begins to improve. However, owing to our features having been normalized, the noise contained in the features is normalized as well. With the $\sigma$ going larger, as a result, our enhanced denoising autoencoder cannot remove noise without keeping useful information.

## 5.2   Comparison of Different Partition Intervals

Unlike [18] regarding the partition interval of ages as 1, precipitation values own a finer ranging interval in our task. And the partition interval decides how many binary classifiers are necessary, and further determines the number of parameters. Therefore, It is crucial to choose a partition interval for ordinal regression in bias correction task. It can be seen from Figure 4(b) that when $\eta$ is set to 0.5, the best partition interval can be obtained. We also notice that the minimum ranging interval cannot obtain the best partition interval. We think that some neighbor rank classes will contain the same training samples when choosing the minimum ranging interval as the partition interval. This situation causes the some samples being classified as positive in these neighbor rank classes and leads to an inaccurate ordinal regression results.

## 6   Conclusions

In this paper, we introduced a novel ordinal boosting regression autoencoder to correct the bias of the European Model precipitation prediction. The ordinal boosting regression autoencoder can extract more robust features from the highly noisy predictive data and correct the biased precipitation value against the maldistribution and a large span of the labels. Experiments on the European Model dataset has indicated that compared with several baseline methods, OBA achieves the best correcting performance and threat scores. The visualization examples reveal OBA has significant correction effect both on the non-precipitation and precipitation situation. In summary, our proposal is effective and practical in the actual application scenario.

In the future, we plan to explore how to utilize finer partition interval to obtain more precise bias correction results. And how to combine the timing information of the predictive weather data with ordinal regression is another important research topic for the bias correction of numerical weather prediction.